\documentclass[10pt,letterpaper,twocolumn]{article}
\usepackage[letterpaper,textwidth=6.875in,textheight=8.875in,centering]{geometry}
\usepackage{times}
\usepackage{amsmath,amssymb}
\usepackage{graphicx}
\usepackage{xcolor,colortbl}
\usepackage{booktabs,tabularx,multirow}
\usepackage{float}
\usepackage[numbers,sort&compress]{natbib}
\usepackage[font=small,labelsep=period]{caption}
\usepackage[most]{tcolorbox}
\usepackage[hidelinks,hypertexnames=false]{hyperref}

\hypersetup{
    pdftitle={InfiMed2: Medical MLLMs from Heterogeneous Evidence and Stability-Aware Explanatory Supervision},
    pdfauthor={Guanghao Zhu, Zeyu Liu, Zhitian Hou, Pengkai Wang, Zhijie Sang, Shuo Cai, Yang Yu, Yuanyi Wang, Yanggan Gu, Congkai Xie, Jianmin Wu, Hongxia Yang}
}

\newenvironment{caseblock}
    {\par\noindent\begin{minipage}{\linewidth}%
     \setlength{\intextsep}{4pt}%
     \setlength{\abovecaptionskip}{4pt}}
    {\end{minipage}\par\medskip}
\newcommand{\casegraphic}[1]{%
    \includegraphics[width=\linewidth,trim=8 26 8 10,clip]{#1}}
\newtcolorbox{promptbox}{
    enhanced,
    breakable,
    colback=black!3,
    colframe=black!35,
    boxrule=0.5pt,
    arc=2pt,
    left=6pt,
    right=6pt,
    top=5pt,
    bottom=5pt,
    before skip=6pt,
    after skip=6pt
}

\title{\textbf{InfiMed2: Medical MLLMs from Heterogeneous Evidence\\ and Stability-Aware Explanatory Supervision}}
\author{
    Guanghao Zhu$^{1}$\thanks{Equal contribution.}\quad
    Zeyu Liu$^{1}$\footnotemark[1]\quad
    Zhitian Hou$^{1}$\footnotemark[1]\quad
    Pengkai Wang$^{1}$\quad
    Zhijie Sang$^{2}$\\[2pt]
    Shuo Cai$^{1}$\quad
    Yang Yu$^{1}$\quad
    Yuanyi Wang$^{1}$\quad
    Yanggan Gu$^{1}$\quad
    Congkai Xie$^{2}$\\[2pt]
    Jianmin Wu$^{1,3}$\quad
    Hongxia Yang$^{1,2,3}$\thanks{Corresponding author: \texttt{hongxia.yang@polyu.edu.hk}.}\\[6pt]
    \small $^{1}$The Hong Kong Polytechnic University\quad
    $^{2}$InfiX.ai\\
    \small $^{3}$Research Institute of Generative Artificial Intelligence
}
\date{}

\begin{document}
\maketitle

\begin{abstract}
Medical multimodal large language models (MLLMs) need both diverse medical evidence and supervision that connects visual findings to medical knowledge. Continued pretraining (CPT) must combine sources whose useful evidence can be obscured by extraction errors, missing figure context, or web noise. In supervised fine-tuning (SFT), short visual question answering (VQA) targets specify the answer but rarely explain which findings support it. To address these problems, we organize medical e-books, biomedical literature, and medical webpages through source-specific curation and combine them with general replay in a 55.68B-token corpus for approximately 55B tokens of warmup--stable CPT. For SFT, stability-aware response regeneration routes questions by reference-answer recovery and uses answer-masked reconstruction with correctness-constrained selection to build explanatory targets. Across five medical multimodal benchmarks, InfiMed2-4B reaches 66.73\% mean accuracy after reinforcement learning with verifiable rewards (RLVR), surpassing Qwen3.5-9B. InfiMed2-27B reaches 73.72\%, the highest mean among the evaluated open-weight models. Our methods for organizing evidence and constructing explanatory supervision, together with the resulting InfiMed2 models, provide a foundation for studying how medical MLLMs learn from visual evidence and use it to explain their answers.
\end{abstract}

\section{Introduction}
\label{sec:introduction}

Medical multimodal large language models (MLLMs) adapt general vision--language capabilities to tasks that require interpreting specialized visual evidence in light of medical knowledge. They are studied for applications ranging from radiology and pathology image interpretation to biomedical question answering and clinical reasoning. Recent work has expanded domain-specific corpora, adapted visual representations to medical images, and strengthened instruction following and reasoning through post-training~\citep{li2023llavamed,chen2024huatuogptvision,sellergren2025medgemma,xu2025lingshu,jiang2025hulumed,shi2026medxiaohe}. These advances broaden what medical MLLMs can do, but leave open how to design the data used to teach them.

Two data-design problems remain. The first is how to organize heterogeneous medical evidence for continued pretraining (CPT). Medical e-books provide systematic knowledge, but parsing errors, duplicate books, and uneven coverage of disciplines and genres complicate large-scale extraction. Biomedical articles describe findings through figures, captions, and passages that refer to those figures. Isolated image--caption pairs provide concentrated alignment signals but omit much of the surrounding evidence, such as the experimental conditions and comparisons needed to interpret a figure. Medical webpages extend the corpus to practical topics, but useful passages can be buried in navigation text, duplicated across pages, or separated from their images. These failure modes require different remedies: a cleaner text extraction does not recover a missing figure reference, and restoring a missing image does not remove redundant web content. The challenge is to curate these sources into a clean CPT corpus without discarding the evidence that makes each source useful.

The second problem is how to construct explanatory supervision for supervised fine-tuning (SFT). Medical visual question answering (VQA) annotations typically give an entity, category, or short judgment~\citep{zhang2023pmcvqa}. These targets identify the expected answer but provide little supervision on which findings to recognize or how to connect them to medical knowledge. Replacing them with a single teacher-generated explanation is not enough: a response can be fluent and detailed while contradicting the reference answer or citing findings that are not visible. The goal is therefore not simply to make answers longer, but to provide more supervision on the connection between visual findings and medical knowledge while maintaining an explicit check on answer correctness.

Therefore, we organize CPT evidence according to how each source presents it. We clean and rebalance e-books by medical subject and genre, retain figure-linked context from biomedical literature, and restore interleaved content from filtered medical webpages. Together with general multimodal replay, these sources form a 55.68B-token corpus. We then draw on each source's high-value subsets: guidelines and textbooks from e-books, medical articles from the web, and clinically relevant figure-linked records from PMC literature. After warmup--stable CPT on the broad corpus, we use a compact mixture enriched in these subsets during learning-rate decay.

For SFT, we construct explanations with stability-aware response regeneration. We route questions by how often a teacher recovers the reference answer across eight samples. In the high-stability branch, we mask the answer and reconstruct the response, then rank only candidates whose reconstructed answers still match the reference. The routing also assigns intermediate-stability questions to a stronger generator and discards questions with low teacher recovery. The resulting response targets supplement short-answer supervision and are mixed with other instruction data by visual modality.

We apply the evidence organization and explanation construction methods to train the InfiMed2 family. Under a common protocol on five medical multimodal benchmarks, InfiMed2-4B reaches 66.73\% mean accuracy after reinforcement learning with verifiable rewards (RLVR), surpassing Qwen3.5-9B. InfiMed2-27B leads our open-weight comparison at 73.72\% after SFT.

Our contributions are summarized as follows:
\begin{itemize}
    \item \textbf{Heterogeneous medical evidence organization.} We curate e-books, biomedical articles, and webpages according to how each source presents evidence. Combining structured medical knowledge, figure-linked passages, and practical web content with general replay yields a 55.68B-token CPT corpus.
    \item \textbf{Explanatory supervision construction.} We develop stability-aware response regeneration with reference-conditioned routing, answer-masked reconstruction, and correctness-constrained selection to construct targets that connect visual findings with medical knowledge.
    \item \textbf{InfiMed2 models.} Our 4B model outperforms Qwen3.5-9B on the five-benchmark mean, and our 27B model achieves the highest mean among the evaluated open-weight models.
\end{itemize}

\section{Related Work}
\label{sec:relatedwork}

\subsection{Medical Multimodal Foundation Models}

Medical multimodal large language models commonly adapt general-purpose vision-language backbones with domain-specific image--text and instruction data. Early systems such as LLaVA-Med~\citep{li2023llavamed} and HuatuoGPT-Vision~\citep{chen2024huatuogptvision} emphasize biomedical alignment and instruction following, while MedGemma~\citep{sellergren2025medgemma} adapts both visual and language components. Recent generalist systems broaden data, modalities, and reasoning supervision, including Lingshu~\citep{xu2025lingshu}, Hulu-Med~\citep{jiang2025hulumed}, and MedXiaoHe~\citep{shi2026medxiaohe}. Despite this progress, source-aware evidence curation and verification of synthetic explanations remain less systematically studied. InfiMed2 addresses these data-design problems across continued pretraining and post-training.

\subsection{Multimodal Continued Pretraining and Medical Data Curation}

Multimodal continued pretraining depends on both coverage and the relation between images and textual evidence. PMC-15M~\citep{zhang2023biomedclip} and BIOMEDICA~\citep{lozano2025biomedica} provide large biomedical figure collections, but literature-derived data often reduce evidence to isolated figure--caption pairs and remain vulnerable to extraction noise. PMC-InterCPT~\citep{zhu2026pmcintercpt} instead reconstructs associations among figures, captions, and their related context. Source diversity, source-specific failure modes, and stage-dependent data value nevertheless remain underexplored. InfiMed2 addresses these limitations by combining complementary medical sources through source-aware processing.

\subsection{Medical Post-Training and Reliable Synthetic Supervision}

Medical post-training uses supervised question answering, instruction following, and reasoning data, sometimes followed by outcome-based reinforcement learning. PMC-VQA~\citep{zhang2023pmcvqa} supplies biomedical VQA supervision, while ReasonMed~\citep{sun2025reasonmed} and recent medical models use synthetic reasoning to strengthen complex problem solving~\citep{xu2025lingshu,shi2026medxiaohe}. Generated explanations, however, may be weakly grounded or inconsistent with verified answers. Self-consistency~\citep{wang2023selfconsistency} and predictive uncertainty~\citep{kuhn2023semanticuncertainty} provide useful reliability signals. Our regeneration pipeline combines these signals with reference-answer constraints to curate explanatory supervision.

\section{Data and Supervision Construction}
\label{sec:data-curation}

\begin{figure*}[t]
    \centering
    \includegraphics[width=\textwidth]{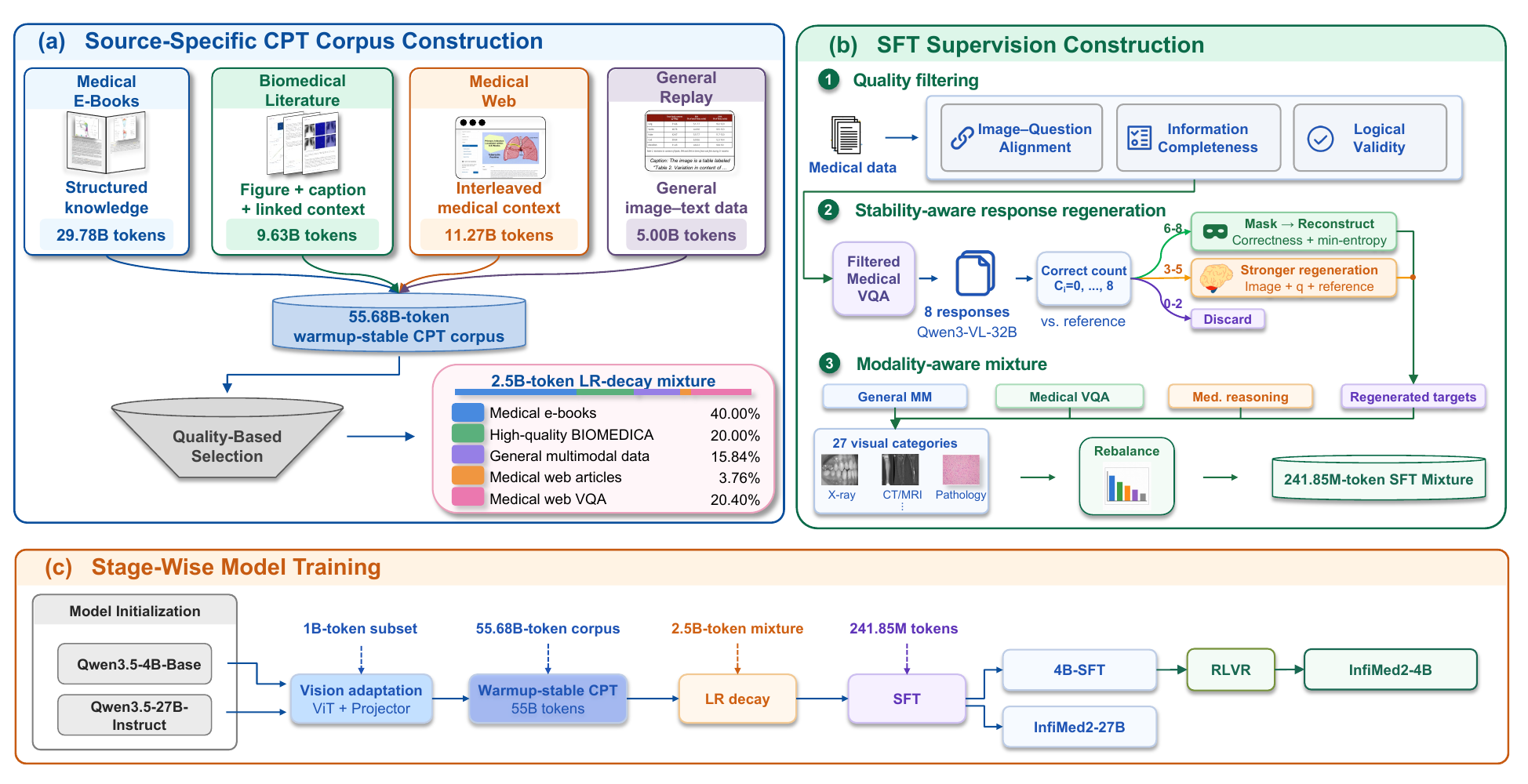}
    \caption{InfiMed2 data construction and training. (a) Four sources form the warmup--stable CPT corpus, and quality-based selection yields the learning-rate-decay mixture. (b) After quality filtering, medical VQA samples are routed by how many of eight teacher responses match the reference answer. Regenerated responses join other supervision in a modality-aware SFT mixture. (c) Training proceeds through vision adaptation, CPT, and SFT. We additionally apply RLVR to the 4B model.}
    \label{fig:infimed2-pipeline}
\end{figure*}

Figure~\ref{fig:infimed2-pipeline} summarizes the construction of the CPT corpus and SFT supervision, and shows how they are used across training stages.

\subsection{Source-Specific CPT Corpus Construction}
\label{sec:cpt-data-curation}

\paragraph{Curation objective and source complementarity.}
We construct the CPT corpus by combining complementary sources of medical knowledge and multimodal evidence, rather than applying a uniform curation strategy to heterogeneous training data: \textbf{(1)} medical e-books provide systematic, long-form exposition; \textbf{(2)} scientific articles connect biomedical figures to peer-reviewed textual evidence; \textbf{(3)} web documents broaden coverage of practical and less formal medical content; and \textbf{(4)} general multimodal data serves as replay for capabilities acquired by the backbone. Because each source exhibits distinct types of noise and distributional bias, we develop source-specific cleaning and sampling procedures for each corpus. After source-specific processing, we obtain a 55.68B-token corpus that is used in its entirety for warmup--stable CPT. It comprises 29.78B tokens from medical e-books, 9.63B from biomedical scientific literature, 11.27B from medical web documents, and 5B from general multimodal data.

\noindent\textbf{Medical e-books.}
We collect approximately 156k medical e-books and convert them into structured Markdown using MinerU~\citep{wang2024mineru}. After book-level deduplication and cleaning, we classify books by medical subject and genre using sampled pages and bibliographic metadata. Subject labels distinguish clinical disciplines from basic medicine, while genre labels distinguish textbooks and guidelines from other material. We retain all textbooks and guidelines, compared with 83.9\% of reference works and 11.4\% of popular-science books. After rebalancing, clinical disciplines account for approximately 75\% of e-book text tokens, up from 55\%, and textbooks account for 27.7\%, up from 13.6\%. The final subset contains 60,433 books and 22.41B text tokens, or 29.78B tokens including visual tokens. Appendix~\ref*{app:ebook-curation} provides the full taxonomy and cleaning rules.

\noindent\textbf{Biomedical scientific literature.}
For biomedical scientific literature, we use 10.11M examples totaling 9.63B tokens from PMC-InterCPT~\citep{zhu2026pmcintercpt}, which reconstructs context-grounded records from BIOMEDICA~\citep{lozano2025biomedica}. The pipeline recovers provenance-linked captions, associates each figure with the context that explicitly references it, and places figures sharing the same context in one interleaved sequence. It then repairs incoherent context, applies medical-relevance and text-quality filtering, and excludes PMC-VQA test examples~\citep{zhang2023pmcvqa}. Unlike isolated image--caption pairs, the resulting corpus preserves the relationships among scientific figures, captions, and figure-linked context, providing supervision for integrating evidence across multiple related elements. Additional construction details are provided in Appendix~\ref*{app:scientific-literature-curation}.

\noindent\textbf{Medical web data.}
Medical web data complements books and scientific articles with broader practical and long-tail medical content. We train a Qwen3-1.7B~\citep{yang2025qwen3} medical-relevance classifier to retrieve a 32.99B-token candidate pool from OmniCorpus-CC~\citep{li2024omnicorpus}, OBELICS~\citep{laurencon2023obelics}, and MINT-1T~\citep{awadalla2024mint1t}, followed by MinHash near-duplicate removal~\citep{broder1997resemblance}. A dedicated model removes navigation, header, and footer text, and we restore interleaving by downloading images referenced by source URLs and rejoining them with their records. We discard documents assigned the lowest score by a three-level quality classifier, then use image categories for balanced resampling. The resulting medical interleaved corpus contains 11.27B tokens.

Relevance, text quality, and image availability are separate checks: a medically relevant page can still contain unusable text or reference an unavailable image. Deduplication reduces the retrieved pool from 22.90M to 16.20M documents, and the quality filter excludes 2.19M records with a zero score or invalid prediction. For image recovery, 22.59M of the 30.02M referenced URLs yield a downloadable image.

On a balanced set of 200 held-out documents, the retrieval classifier reaches 98.5\% accuracy. The quality classifier achieves 90.3\% precision on the discarded score-0 class, measured on a separate set of 255 documents. These figures evaluate individual filters rather than the end-to-end quality of the web corpus.

\noindent\textbf{General multimodal replay.}
Medical specialization can weaken broad visual and linguistic capabilities if training excludes general-domain evidence. We therefore include 5B tokens from LLaVA-OneVision-1.5-Mid-Training-85M~\citep{an2025llavaonevision} as general multimodal replay during warmup--stable CPT. The source dataset contains 85M curated and concept-balanced examples, and its associated models demonstrate strong performance across a broad benchmark suite. We convert a 30M-example intermediate pool containing 19.89B tokens and randomly sample the 5B-token training subset. We keep this component smaller than the combined medical sources so that it regularizes domain adaptation without diluting the medical focus of the mixture.

% \begin{table}[t]
%     \centering
%     \caption{Composition of the warmup--stable CPT corpus. Each source is processed with a source-specific pipeline before token-budgeted mixture construction.}
%     \label{tab:cpt-warmup-data}
%     \small
%     \begin{tabular}{l r r p{5.2cm}}
%         \toprule
%         \textbf{Source}         & \textbf{Tokens} & \textbf{Share} & \textbf{Primary contribution}                               \\
%         \midrule
%         Medical e-books         & 29.78B          & 53.48\%        & Clinically rebalanced long-form knowledge                   \\
%         PMC-InterCPT            & 9.63B           & 17.30\%        & Context-grounded biomedical scientific evidence             \\
%         Medical web data        & 11.27B          & 20.24\%        & Broad practical coverage and restored interleaved documents \\
%         General multimodal data & 5.00B           & 8.98\%         & General-capability replay                                   \\
%         \midrule
%         Total                   & 55.68B          & 100.00\%       & Warmup--stable CPT corpus                                   \\
%         \bottomrule
%     \end{tabular}
% \end{table}

\subsection{Fine-Grained Quality Filtering}
\label{sec:data_pipeline}
\label{sec:quality_filtering}

We collect post-training data from heterogeneous medical and general multimodal sources. Directly combining these sources introduces three practical issues. First, some samples contain semantic inconsistencies that metadata cannot reliably identify, including mismatched image modalities, missing anatomical evidence, incomplete problem conditions, or responses not properly grounded in the visual input. Second, many medical VQA datasets provide only short-form answers, which are sufficient for evaluating answer correctness but provide limited supervision for learning informative multimodal responses. Third, the resulting corpus is highly imbalanced in visual content.

As shown in Figure~\ref{fig:infimed2-pipeline}(b), we first apply fine-grained quality filtering to candidate medical data. We then regenerate responses for retained medical visual question answering (VQA) samples. The resulting responses are combined with medical VQA, medical reasoning, and general multimodal instruction data before modality-aware sampling.

We first assess the validity of each candidate sample by jointly considering its visual and textual components. We define a fine-grained annotation scheme along three dimensions: image--question alignment, information completeness, and logical validity. The annotation covers common failure cases such as image-modality mismatch, missing anatomical regions, and inconsistent question--answer pairs.

Based on this annotation scheme, we construct a quality-control dataset and train a 2B parameter evaluator to assess candidate samples. We use a three-level score, where Score~0 indicates a critical defect, Score~1 denotes a limited but non-fatal issue, and Score~2 denotes a valid sample. The evaluator removes Score~0 samples before response construction and mixture formation. The complete quality assessment protocol is provided in Appendix~\ref*{app:posttraining-quality} and Table~\ref*{tab:quality_criteria}.

\subsection{Stability-Aware Response Regeneration}
\label{sec:answer_regeneration}

After filtering, we refine the response supervision of retained medical VQA data. Existing medical VQA datasets often provide only an entity, category label, or short judgment and therefore provide limited supervision for integrating visual evidence with medical knowledge. Our regeneration procedure produces informative explanations while preserving the reference answer.

\paragraph{Multi-sample response generation and stability estimation.}
Let $x_i=(I_i,q_i,a_i^*)$ denote a medical VQA instance, where $I_i$ denotes the medical image, $q_i$ the question, and $a_i^*$ the reference answer. We first employ Qwen3-VL-32B-Instruct~\citep{bai2025qwen3vl} parameterized by $\theta_T$ as a teacher model to independently sample $K=8$ candidate responses:
\begin{equation}
    r_i^{(k)} \sim p_{\theta_T}(r\mid I_i,q_i),
    \qquad k=1,\ldots,K.
    \label{eq:teacher_sampling}
\end{equation}

Inspired by repeated sampling in self-consistency decoding~\citep{wang2023selfconsistency}, we measure reference-conditioned answer stability: agreement with the reference answer, rather than agreement among teacher responses.

Let $\hat a_i^{(k)}$ be the final answer extracted from $r_i^{(k)}$, and let $\mathcal M$ denote the dataset-specific answer-matching function. We count the correct responses and normalize this count as
\begin{equation}
    C_i=\sum_{k=1}^{K}\mathbb I\!\left[\mathcal M(\hat a_i^{(k)},a_i^*)=1\right],
    \qquad
    S_i=\frac{C_i}{K}.
    \label{eq:response_stability}
\end{equation}
Here, $\mathbb I[\cdot]$ is the indicator function, $C_i$ is the number of correct responses, and $S_i$ measures how consistently the teacher recovers the reference answer. As $K$ is fixed, we use $C_i$ for routing.

\paragraph{Answer-masked reconstruction and candidate selection.}
For high-stability samples with $6\leq C_i\leq8$, we apply answer masking to correct teacher responses, yielding $\tilde r_i^{(k)}=\operatorname{MaskAnswer}(r_i^{(k)},\hat a_i^{(k)})$. Qwen3-VL-4B-Instruct~\citep{bai2025qwen3vl}, parameterized by $\theta_R$, then reconstructs the response from the image, question, and masked explanation:
\begin{equation}
    \bar r_i^{(k)} \sim p_{\theta_R}(r\mid I_i,q_i,\tilde r_i^{(k)}).
    \label{eq:masked_reconstruction}
\end{equation}
Here, $\bar r_i^{(k)}$ denotes the reconstructed response. We form $\mathcal R_i^+$ from reconstructions whose extracted answers match $a_i^*$ and whose originating teacher responses are also correct.

Entropy-based quantities provide a natural measure of uncertainty in probabilistic language generation~\citep{kuhn2023semanticuncertainty}. In our setting, reference-answer consistency is first enforced as a hard constraint, after which entropy is used only to rank the remaining valid candidates. The ranking rule is written in terms of mean token entropy:
\begin{equation}
    \mathcal H(\bar r_i^{(k)})
    =\frac{1}{T_k}\sum_{t=1}^{T_k}
    \left[-\sum_{v\in\mathcal V}p_t(v)\log p_t(v)\right].
    \label{eq:mean_entropy}
\end{equation}
Here, $T_k$ is the length of the scored token span, $\mathcal V$ is the output vocabulary, and $p_t(v)$ is the reconstruction model's probability of token $v$ at position $t$. Among the valid candidates, we select
\begin{equation}
    r_i^*=\arg\min_{r\in\mathcal R_i^+}\mathcal H(r).
    \label{eq:response_selection}
\end{equation}
If $\mathcal R_i^+=\emptyset$, we discard the sample.

\paragraph{Escalated regeneration and routing.}
For intermediate-stability samples with $3\leq C_i\leq5$, Gemini 3.1 Pro~\citep{gemini31pro2026} generates a new response conditioned on $(I_i,q_i,a_i^*)$. Samples with $C_i<3$ are discarded. The routing thresholds reflect teacher answer recovery, not an independent assessment of sample quality.

Of the 176,948 input questions, 121,544 have at least one answer-consistent teacher candidate. The pipeline exports 108,293 long-form examples; 106,183 enter the final SFT mixture. Among 704,046 reported answer-masked reconstructions, 662,503 (94.10\%) recover the reference answer. The reconstructions average 11.62 assistant tokens, compared with 418.99 for the exported explanations. The short reconstruction probes answer recovery, while the longer response supplies the supervision target. Answer recovery does not certify the explanation's factual accuracy or visual grounding.

\subsection{Visual-Modality-Aware SFT Mixture}
\label{sec:modality_aware_mixture}

After filtering and regeneration, we combine the regenerated examples with general multimodal instructions, medical VQA, and medical reasoning data. A trained classifier annotates samples with 27 visual categories spanning radiology, pathology, plots, diagrams, and other visual content. Category-level sampling reduces domination by frequent modalities while preserving clinically relevant long-tail content.

The resulting SFT mixture contains 296,725 examples and 241.85M tokens. Stability-aware supervision contributes 106,183 examples and 108.32M tokens, or 44.79\% of the token budget. The remaining tokens come from medical reasoning (22.69\%), medical VQA (21.77\%), and general multimodal instructions (10.76\%). Thus the regenerated explanations account for a larger share of the token budget than of the examples, without replacing the other forms of supervision. Appendix~\ref*{app:sft-mixture} reports the visual and textual token counts by category.

\section{Model Training}
\label{sec:model-training}

The 4B and 27B training runs are initialized from Qwen3.5-4B-Base and the instruction-tuned Qwen3.5-27B checkpoint, respectively. Figure~\ref{fig:infimed2-pipeline}(c) shows the stage order: vision-encoder adaptation, warmup--stable CPT, learning-rate decay, and SFT. Table~\ref{tab:checkpoint-rlvr-ablation} distinguishes the evaluated checkpoints. We apply RLVR only to the SFT checkpoint of the selected 4B variant with 30B warmup--stable tokens.

\subsection{Stage-Specific Continued Pretraining}
\label{sec:cpt-model-training}
\label{sec:stage-specific-cpt-data}

% Queue the main results table before the first experiments page.
\begin{table*}[t]
\centering
\caption{Medical multimodal accuracy (\%) under a common evaluation protocol. The shaded InfiMed2-4B and InfiMed2-27B rows report results after SFT+RLVR and SFT, respectively. Bold and underlined entries denote the best and second-best scores.}
\label{tab:medical-main-results}
\footnotesize
\setlength{\tabcolsep}{3pt}
\renewcommand{\arraystretch}{1.08}
\begin{tabularx}{\textwidth}{l*{5}{>{\centering\arraybackslash}X}>{\raggedleft\arraybackslash}X}
\toprule
\textbf{Model} & \shortstack{\textbf{MMMU}\\\textbf{Med.}} & \shortstack{\textbf{MMMU-Pro}\\\textbf{Med.-10}} & \textbf{MedXQA} & \shortstack{\textbf{PMC-VQA}\\\textbf{clean}} & \shortstack{\textbf{OmniMed}\\\textbf{VQA}} & \textbf{Avg.} \\
\midrule
\multicolumn{7}{c}{\emph{General-purpose models}} \\
\midrule
Gemini-3-Pro & 81.84 & \textbf{74.13} & \textbf{74.85} & \textbf{70.40} & 84.55 & \textbf{77.15} \\
Gemini-3-Flash & 81.51 & 70.63 & 68.00 & \underline{69.75} & 84.45 & \underline{74.87} \\
Claude-Opus-4.7 & \underline{82.47} & 70.63 & 69.40 & 67.15 & 81.60 & 74.25 \\
GPT-5 & \textbf{83.39} & 70.90 & \underline{71.70} & 67.30 & 76.40 & 73.94 \\
GLM-5V-Turbo & 73.91 & 64.69 & 53.25 & 66.55 & 81.95 & 68.07 \\
Inkling & 73.17 & 60.49 & 56.00 & 57.80 & 69.05 & 63.30 \\
MiMo-v2.5 & 73.00 & 56.65 & 50.20 & 53.50 & 70.55 & 60.78 \\
Qwen3.5-4B & 68.66 & 54.89 & 34.75 & 60.15 & 82.50 & 60.19 \\
Qwen3.5-9B & 76.31 & 58.39 & 40.35 & 62.60 & 85.91 & 64.71 \\
Qwen3.5-27B & 74.70 & 60.48 & 47.05 & 64.30 & 90.60 & 67.43 \\
Gemma 4-31B & 79.16 & 66.43 & 54.55 & 66.00 & 80.15 & 69.26 \\
\midrule
\multicolumn{7}{c}{\emph{Medical-domain models}} \\
\midrule
MedGemma-1.5-4B-IT & 47.26 & 30.42 & 28.40 & 47.75 & 69.38 & 44.64 \\
Lingshu-32B & 62.30 & 41.26 & 30.90 & 57.90 & 83.40 & 55.15 \\
Hulu-Med-32B & 60.84 & 40.55 & 34.10 & 64.55 & 84.90 & 56.98 \\
\midrule
\multicolumn{7}{c}{\emph{Selected InfiMed2 models (ours)}} \\
\midrule
\rowcolor[gray]{0.93} InfiMed2-4B & 72.43 & 58.39 & 46.25 & 65.80 & \underline{90.79} & 66.73 \\
\addlinespace[2pt]
\rowcolor[gray]{0.93} InfiMed2-27B & 77.85 & \underline{70.98} & 59.40 & 67.50 & \textbf{92.93} & 73.72 \\
\bottomrule
\end{tabularx}
\end{table*}

\paragraph{Vision-encoder adaptation.}
We construct a visually concentrated subset from PMC-InterCPT by retaining the first valid image and its caption in each eligible record. Deterministic token-budgeted sampling yields 1.71M pairs totaling 1B tokens, of which 73.7\% are visual tokens. We train the vision encoder and multimodal projector for one epoch on this subset while keeping the language model frozen. The goal is to align medical images with their captions before introducing context-rich interleaved records.

\paragraph{Warmup--stable continued pretraining.}
We then train all model components for one epoch on the 55.68B-token corpus, completing approximately 55B tokens of warmup--stable CPT. The mixture exposes the model to diverse medical sources while retaining general replay.

\paragraph{Learning-rate decay.}
Finally, we train all model components for one epoch on the high-quality 2.50B-token mixture. Medical guidelines and textbooks from the e-book corpus account for 40\% of the tokens, high-quality medical samples from PMC-InterCPT for 20\%, and general multimodal data for 15.84\%. The remaining 24.16\% consists of medical web articles (3.76\%) and medical exercises and selected VQA examples (20.40\%).

All CPT stages use AdamW with a global batch size of 256. The maximum sequence length is 8,192 for vision-encoder adaptation and 12,288 for the two subsequent stages. Vision adaptation uses 3\% warmup to $2\times10^{-6}$ followed by cosine decay to $2\times10^{-7}$. Warmup--stable CPT reaches $1\times10^{-5}$ after 3\% warmup and holds it constant, while the final stage decays it from $1\times10^{-5}$ to $2\times10^{-6}$. Packed sequences preserve sample boundaries through position IDs. Image placeholders and delimiters are excluded from the language-model loss.

% \begin{table}[t]
% \centering
% \caption{Data mixture for the CPT learning-rate decay stage.}
% \label{tab:cpt-decay-data}
% \small
% \begin{tabular}{l r r}
% \toprule
% \textbf{Data source} & \textbf{Tokens} & \textbf{Share} \\
% \midrule
% Curated medical e-books & 1.000B & 40.00\% \\
% BIOMEDICA & 0.500B & 20.00\% \\
% General multimodal data & 0.396B & 15.84\% \\
% Medical web articles & 0.094B & 3.76\% \\
% Medical web exercises and selected VQA data & 0.510B & 20.40\% \\
% \midrule
% Total & 2.500B & 100.00\% \\
% \bottomrule
% \end{tabular}
% \end{table}

\subsection{Supervised Fine-Tuning}
\label{sec:sft-model-training}

Starting from the post-decay checkpoint, we train all model components for five epochs on the 241.85M-token SFT mixture (Section~\ref{sec:modality_aware_mixture}). The token count describes the mixture, not the total exposure over five epochs. We use AdamW with a global batch size of 64 and a maximum sequence length of 9,000. After 10\% warmup, the learning rates peak at $2\times10^{-6}$ for the language model and projector and $1\times10^{-6}$ for the vision encoder, followed by cosine decay to zero.

\subsection{RL with Verifiable Rewards}
\label{sec:rlvr}
Starting from the selected 4B SFT checkpoint, we perform RLVR using a resource of 21,529 examples, comprising 19,529 medical VQA examples and 2,000 text-only USMLE questions drawn from public resources and internally curated data~\citep{zhang2023pmcvqa,huang2026medvlsynther,li2026gmaivl,hu2024omnimedvqa,mei2022radimagenet,jin2021medqa}. This total includes the train and test partitions in Appendix~\ref*{app:rlvr-data}. We train for three epochs with a learning rate of $1\times10^{-6}$ and a global batch size of 128. The policy samples eight rollouts per prompt at temperature 1. We weight format compliance and answer accuracy in the reward at $1{:}9$.

We estimate difficulty by the number of correct rollouts out of eight and prioritize examples with 3--5 correct, where reward remains attainable but improvement is still possible. We downsample near-unsolved and saturated examples but keep a smaller easy subset to help preserve capabilities learned during SFT. Full optimization and implementation details are provided in Appendix~\ref*{app:training-details}.

\section{Experiments}
\label{sec:experiments}

\subsection{Evaluation Protocol and Baselines}
\label{sec:experimental-setup}

\paragraph{Evaluation benchmarks.}
We evaluate medical multimodal understanding on five benchmarks: the Health and Medicine test subset of MMMU~\citep{yue2024mmmu}, the corresponding 10-option subset of MMMU-Pro~\citep{yue2025mmmupro}, MedXpertQA-MM~\citep{zuo2025medxpertqa}, PMC-VQA-clean~\citep{zhang2023pmcvqa}, and OmniMedVQA~\citep{hu2024omnimedvqa}. We report accuracy on each benchmark and the unweighted mean across the five benchmarks.

\paragraph{Baselines and evaluation protocol.}
We compare against general-purpose and medical multimodal models, including the Qwen3.5 backbones and medical models such as MedGemma~\citep{sellergren2025medgemma}, Lingshu~\citep{xu2025lingshu}, and Hulu-Med~\citep{jiang2025hulumed}. We also train SFT-only controls from Qwen3.5-4B-Base and the instruction-tuned Qwen3.5-27B checkpoint, using the same instruction data as their CPT-initialized counterparts.

All scores use a shared evaluation protocol rather than published leaderboard numbers. We disable native thinking and cap generation at 4,096 tokens. OmniMedVQA uses a direct-answer prompt, while the other four benchmarks use a reasoning prompt. Within each benchmark, all models receive the same questions and prompts, and we score the option letter inside the answer tags by exact match. Appendix~\ref*{app:evaluation-details} provides the prompt templates.

\subsection{Main Results}
\label{sec:main-results}

\begin{figure*}[t]
\centering
\includegraphics[width=\textwidth]{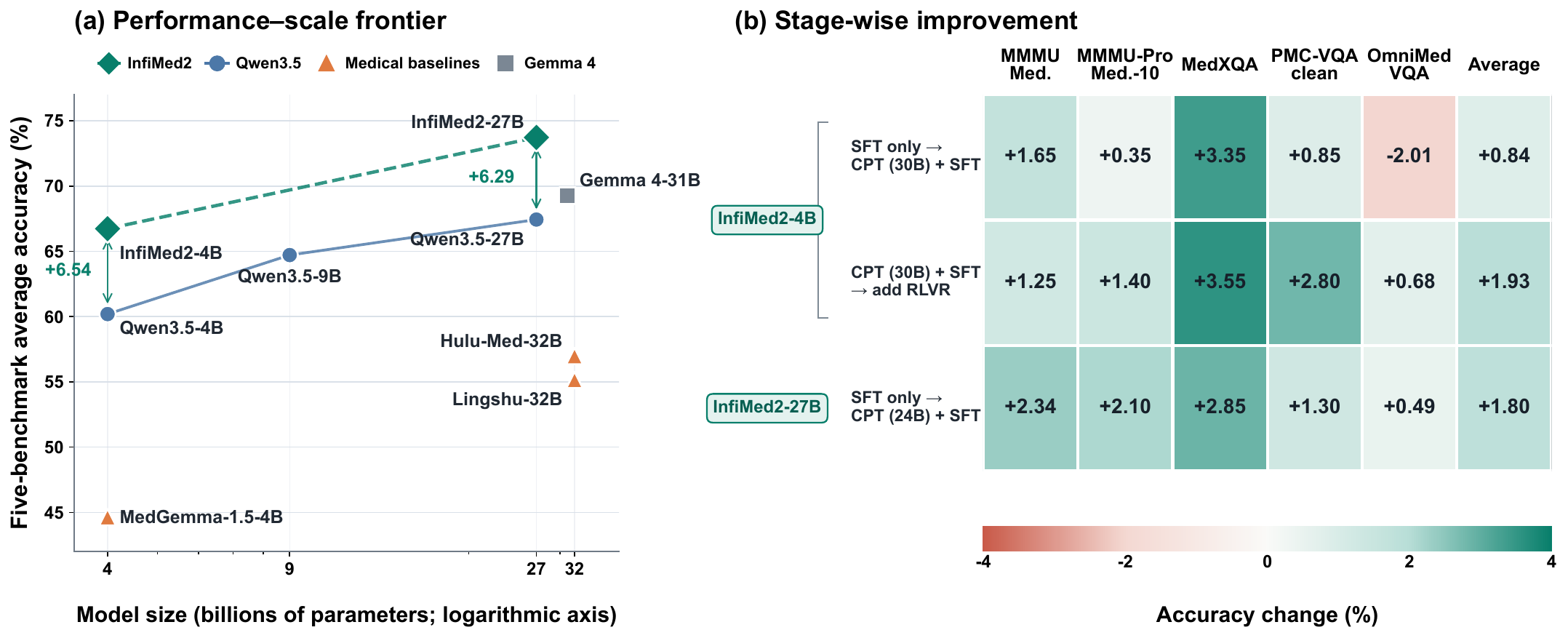}
\caption{Parameter efficiency and stage-wise gains. (a) Five-benchmark average accuracy versus model size for open-weight models with reported parameter counts. (b) Benchmark-wise score changes after adding CPT or RLVR to the corresponding matched configuration.}
\label{fig:performance-stage-analysis}
\end{figure*}

\begin{table*}[t]
\centering
\caption{CPT checkpoint selection and RLVR ablation. CPT columns report token budgets.}
\label{tab:checkpoint-rlvr-ablation}
\footnotesize
\setlength{\tabcolsep}{1pt}
\renewcommand{\arraystretch}{1.00}
\begin{tabular*}{\textwidth}{@{\extracolsep{\fill}}lcccccccccr@{}}
\toprule
\textbf{Parameters} & \shortstack{\textbf{Vision-encoder}\\\textbf{adaptation}} & \shortstack{\textbf{Warmup--}\\\textbf{stable CPT}} & \shortstack{\textbf{Learning-rate}\\\textbf{decay}} & \shortstack{\textbf{Post-}\\\textbf{training}} & \shortstack{\textbf{MMMU}\\\textbf{Med.}} & \shortstack{\textbf{MMMU-Pro}\\\textbf{Med.-10}} & \textbf{MedXQA} & \shortstack{\textbf{PMC-VQA}\\\textbf{clean}} & \textbf{OmniMedVQA} & \textbf{Avg.} \\
\midrule
\multirow{4}{*}{4B} & -- & -- & -- & SFT & 69.53 & 56.64 & 39.35 & 62.15 & \textbf{92.12} & 63.96 \\
 & 1B & 30B & 2.5B & SFT & \underline{71.18} & \underline{56.99} & 42.70 & 63.00 & 90.11 & \underline{64.80} \\
 & 1B & 55B & 2.5B & SFT & 70.55 & 55.94 & \underline{43.05} & \underline{63.90} & 90.04 & 64.69 \\
 & 1B & 30B & 2.5B & SFT+RLVR & \textbf{72.43} & \textbf{58.39} & \textbf{46.25} & \textbf{65.80} & \underline{90.79} & \textbf{66.73} \\
\midrule
\multirow{4}{*}{27B} & -- & -- & -- & SFT & 75.51 & \underline{68.88} & 56.55 & 66.20 & 92.44 & 71.92 \\
 & 1B & 8B & 2.5B & SFT & \underline{78.88} & 66.78 & 57.45 & 66.20 & \underline{92.69} & 72.40 \\
 & 1B & 24B & 2.5B & SFT & 77.85 & \textbf{70.98} & \underline{59.40} & \underline{67.50} & \textbf{92.93} & \textbf{73.72} \\
 & 1B & 55B & 2.5B & SFT & \textbf{79.97} & 67.48 & \textbf{60.05} & \textbf{68.60} & 91.51 & \underline{73.51} \\
\bottomrule
\end{tabular*}
\end{table*}

Table~\ref{tab:medical-main-results} compares the selected InfiMed2 models with general-purpose and medical baselines. After RLVR, InfiMed2-4B reaches 66.73\% average accuracy, outperforming Qwen3.5-4B by 6.54 percentage points and the larger Qwen3.5-9B by 2.02 points. InfiMed2-27B achieves the highest average among the evaluated open-weight models at 73.72\%, trailing GPT-5 and Claude-Opus-4.7 by only 0.22 and 0.53 points, respectively. Figure~\ref{fig:performance-stage-analysis}(a) shows that the evaluated InfiMed2 checkpoints outperform the corresponding Qwen3.5 checkpoints at both sizes.

\subsection{Contributions of CPT and RLVR}
\label{sec:ablation-studies}

% Queue the remaining ablations for the next page of double-column floats.
\begin{table*}[t]
\centering
\caption{Matched lightweight-SFT ablation of response targets.}
\label{tab:response-regeneration-ablation}
\footnotesize
\setlength{\tabcolsep}{2.5pt}
\renewcommand{\arraystretch}{1.00}
\begin{tabular*}{\textwidth}{@{\extracolsep{\fill}}lcccccr@{}}
\toprule
\textbf{SFT target} & \shortstack{\textbf{MMMU}\\\textbf{Med.}} & \shortstack{\textbf{MMMU-Pro}\\\textbf{Med.-10}} & \textbf{MedXQA} & \shortstack{\textbf{PMC-VQA}\\\textbf{clean}} & \textbf{OmniMedVQA} & \textbf{Avg.} \\
\midrule
Answer only & 58.56 & 40.21 & 31.25 & \textbf{65.70} & \textbf{83.19} & 55.78 \\
Random generated response & 68.04 & \textbf{52.10} & \underline{34.85} & 59.25 & 75.60 & 57.97 \\
Random correct reconstruction & \underline{68.84} & 51.40 & 34.15 & \underline{59.75} & 76.40 & \underline{58.11} \\
Stability-aware selected response & \textbf{69.81} & \underline{51.75} & \textbf{35.00} & 59.20 & \underline{77.00} & \textbf{58.55} \\
\bottomrule
\end{tabular*}
\end{table*}

\begin{table*}[t]
\centering
\caption{Matched 100M-token decay-mixture ablation.}
\label{tab:decay-mixture-ablation}
\footnotesize
\setlength{\tabcolsep}{2.5pt}
\renewcommand{\arraystretch}{1.00}
\begin{tabular*}{\textwidth}{@{\extracolsep{\fill}}lcccccr@{}}
\toprule
\textbf{Decay data} & \shortstack{\textbf{MMMU}\\\textbf{Med.}} & \shortstack{\textbf{MMMU-Pro}\\\textbf{Med.-10}} & \textbf{MedXQA} & \shortstack{\textbf{PMC-VQA}\\\textbf{clean}} & \textbf{OmniMedVQA} & \textbf{Avg.} \\
\midrule
Broad CPT mixture (control) & 66.67 & 51.75 & 33.00 & 57.95 & 76.83 & 57.24 \\
High-quality decay mixture & \textbf{67.75} & \textbf{58.04} & \textbf{34.25} & \textbf{59.15} & \textbf{77.11} & \textbf{59.26} \\
\bottomrule
\end{tabular*}
\end{table*}

Under matched SFT, the combined CPT stages improve mean accuracy over SFT-only at both model sizes. Among the tested budgets, the highest CPT+SFT means are at 30B warmup--stable tokens for the 4B model and 24B for the 27B model; both completed 55B configurations also outperform SFT-only on average. At 27B, the 55B checkpoint improves three of five benchmarks over 24B but scores lower on MMMU-Pro-Medical-10 and OmniMedVQA, leaving its mean 0.21 points lower. RLVR then improves all five benchmarks for the selected 4B checkpoint, with the largest gains on MedXpertQA-MM and PMC-VQA-clean. Figure~\ref{fig:performance-stage-analysis}(b) summarizes these matched stage-wise gains, with parenthesized values denoting warmup--stable CPT tokens.

The matched comparisons reveal differences that the mean alone hides. At 4B, adding CPT improves MedXpertQA-MM by 3.35 points but lowers OmniMedVQA by 2.01 points; the subsequent RLVR stage recovers 0.68 points on OmniMedVQA without closing that gap. At 27B, the selected CPT checkpoint improves both benchmarks relative to SFT-only. The data therefore support gains in aggregate, while the effect on broad medical VQA depends on the evaluated model and checkpoint.

\subsection{Response Construction Ablation}
\label{sec:response-regeneration-ablation}

Using the same 4B checkpoint and 20k PMC-VQA examples, we compare four response targets (Table~\ref{tab:response-regeneration-ablation}). Generated explanations improve reasoning-oriented benchmarks over short answers, while short-answer targets remain stronger on PMC-VQA-clean and OmniMedVQA. To separate the ranking step from answer correctness, we add a control that randomly selects an answer-correct reconstruction in place of minimum-entropy selection. The latter improves the mean from 58.11\% to 58.55\% (+0.44 points) and scores higher on four of five benchmarks, with PMC-VQA-clean the exception. The ranking rule uses reconstruction uncertainty to distinguish candidates that all pass the same answer check, preferring those with more concentrated token distributions.

\subsection{Controlled Evaluation of the Decay Mixture}
\label{sec:decay-mixture-ablation}

Both branches start from the same 4B checkpoint after 55B warmup--stable CPT and before decay. The control retains the broad CPT mixture; the proposed condition uses the high-quality decay mixture in Section~\ref{sec:cpt-model-training}. Both use approximately 100M multimodal tokens and cosine decay without warmup, from a learning rate of $1\times10^{-5}$ to $2\times10^{-6}$. Both keep the vision encoder unfrozen, then undergo five epochs of SFT on the same 20k PMC-VQA explanatory targets.

Under this fixed schedule, the high-quality mixture raises mean accuracy from 57.24\% to 59.26\% (+2.02 percentage points; Table~\ref{tab:decay-mixture-ablation}). It improves all five benchmarks, with the largest gain on MMMU-Pro-Medical-10 (+6.29 points) and the smallest on OmniMedVQA (+0.28 points). The other three benchmarks each gain more than one point.

The decay mixture is not simply a smaller sample of the warmup--stable corpus. It restricts the e-book component to medical textbooks and guidelines (40\%) and selects high-quality medical records from PMC-InterCPT (20\%). The broad mixture has no separate quota for medical exercises or VQA, whereas the decay mixture allocates 20.40\% to that category and retains 15.84\% for general replay. This shifts the remaining training budget toward clinical sources and task-specific practice without removing general multimodal data.

The gains support choosing both the sources and their proportions for decay rather than keeping the broad mixture unchanged. Because these choices also change the share of visual tokens, the comparison tests them jointly. It does not identify a single contributor, an optimal ratio, or the gain at the full 2.50B-token budget.

\subsection{Discussion}
\label{sec:discussion-limitations}

Neither longer pretraining nor longer answers consistently improve every task. The preferred checkpoint depends on which capabilities matter downstream, and explanatory targets complement rather than replace concise VQA supervision. Final-answer accuracy also leaves explanation quality unresolved: a model can select the correct option for the wrong medical reason or describe the right anatomy while pointing to the wrong label. Expert assessment of visual grounding and medical reasoning would complement these benchmark scores.

\section{Conclusion}
\label{sec:conclusion}

We present InfiMed2, a family of medical MLLMs trained by organizing heterogeneous medical evidence for CPT and constructing reference-answer-constrained explanatory supervision for SFT. Across five medical multimodal benchmarks, InfiMed2-4B surpasses Qwen3.5-9B, and InfiMed2-27B achieves the highest mean among the evaluated open-weight models. In the matched 100M-token ablation, the curated decay mixture improves mean accuracy by 2.02 points over the broad mixture under the same training schedule. The response-target ablation shows that explanations help on reasoning-oriented tasks, while concise answers remain useful for visual question answering. These findings guide how researchers curate medical evidence and teach models to explain how visual findings support their answers to medical questions.

\section*{LLM Usage Statement}

We used OpenAI ChatGPT and Codex to assist with literature search, code development, and language polishing.

\section*{Acknowledgments}

This work was supported by the Hong Kong RGC (TRS: T41-517/25-N; GRF: 15228325), ITC RAISe+ (RAI/24/1/086A), and the Research Institute for Generative Artificial Intelligence at PolyU.

{\small
\bibliographystyle{ieeenat_fullname}
\bibliography{references}
}

\onecolumn
\appendix
\setcounter{table}{0}
\setcounter{figure}{0}
\renewcommand{\thetable}{S\arabic{table}}
\renewcommand{\thefigure}{S\arabic{figure}}
\raggedbottom
\section{Additional Data Curation Details}
\label{app:data-curation}

\subsection{Medical E-Book Processing and Sampling}
\label{app:ebook-curation}

We collect 156,498 medical e-books across six source formats, of which 156,051 are successfully converted into structured Markdown with MinerU~\citep{wang2024mineru}, corresponding to a 99.7\% file-level conversion rate. Before sampling, we construct a content signature for each parsed book by concatenating its ten longest paragraphs, each containing at least 50 characters. Global exact matching identifies 3,025 duplicate books, or 1.93\% of the inputs to this stage. We further remove parsing failures, empty outputs, samples dominated by repeated characters, and records containing only a cover page. After cleaning and deduplication, 152,987 books remain, containing 45.58B text tokens.

For classification, the model receives sampled pages together with a small set of bibliographic metadata and predicts a constrained major subject, an open-ended fine-grained subject, and one of seven medical genres: monograph, popular science, reference/atlas/quick-reference work, textbook, research literature/report, supplementary teaching material, and guideline/standard. The subject taxonomy follows a medical-first rule derived from the R categories of the Chinese Library Classification. Content concerning human structure or function, health, disease, diagnosis, treatment, or clinical practice is assigned to the most specific applicable medical category. Biology is reserved for content centered on non-human biological systems, while the non-medical category is used only when the material is unrelated to medicine or biology.

The main paper reports changes for clinical disciplines and textbooks. In addition, the text-token share of basic medicine falls from 21.3\% to 9.2\%, the share of medical reference/atlas/quick-reference works increases from 11.1\% to 17.9\%, and the share of medical popular science decreases from 5.6\% to 1.3\%. We retain 7.1\% of medical supplementary teaching materials.

\subsection{Biomedical Scientific Literature}
\label{app:scientific-literature-curation}

PMC-InterCPT~\citep{zhu2026pmcintercpt} first recovers provenance-linked captions from PMC XML and normalizes the source text while preserving figure-reference anchors. It constructs interleaved sequences by attaching figure-linked context only to figures that the context explicitly references. Figures referenced by the same context remain in a shared sequence rather than being paired with duplicated context. A coherence-repair step removes discontinuous context and prunes images that are no longer supported by the retained text. Separate classifiers then filter records by medical relevance and textual quality, and examples overlapping the PMC-VQA test set are excluded~\citep{zhang2023pmcvqa}. Evidence-aware allocation assigns 45\% of the final token budget to biomedical visual evidence, 30\% to quantitative and tabular evidence, 20\% to mechanism and structure evidence, and 5\% to auxiliary evidence.

\subsection{Medical Web Curation Diagnostics}
\label{app:web-curation}

Across MINT-1T, OBELICS, and OmniCorpus-CC, the retrieved candidate pool contains approximately 33.00B text tokens before MinHash deduplication and 23.29B after it. The main paper reports the document and image-URL counts at each stage.

The medical-retrieval classifier is trained on 4,024 annotated examples. The three-level quality classifier is trained on 2,058 examples and obtains 72.5\% accuracy and 73.4\% macro-F1 on 255 held-out documents. The border-text removal model is trained on 4,619 webpages and achieves 82.1\% line-level accuracy, 77.9\% precision, and 83.1\% recall over 4,923 annotated lines from 256 held-out pages.

\subsection{Stage-Specific CPT Mixtures}
\label{app:cpt-mixtures}

Vision-encoder adaptation uses a 1B-token subset of PMC-InterCPT. Each retained record is reduced to one biomedical image and its associated caption. Deterministic token-budget sampling selects 1,712,897 examples containing 262.50M text tokens and 737.50M visual tokens. The learning-rate decay stage instead uses the compact mixture in Table~\ref{tab:decay-mixture}. It emphasizes curated medical sources while retaining a controlled amount of general replay.

\begin{table}[H]
    \centering
    \caption{Composition of the 2.5B-token mixture used during learning-rate decay.}
    \label{tab:decay-mixture}
    \small
    \renewcommand{\arraystretch}{1.08}
    \begin{tabular*}{\textwidth}{@{\extracolsep{\fill}}lrr@{}}
        \toprule
        \textbf{Source} & \textbf{Tokens} & \textbf{Share} \\
        \midrule
        Medical e-books (guidelines and textbooks) & 1.000B & 40.00\% \\
        High-quality medical samples from PMC-InterCPT & 0.500B & 20.00\% \\
        General multimodal replay & 0.396B & 15.84\% \\
        Medical web articles & 0.094B & 3.76\% \\
        Medical web exercises and selected VQA & 0.510B & 20.40\% \\
        \midrule
        Total & 2.500B & 100.00\% \\
        \bottomrule
    \end{tabular*}
\end{table}

\section{Additional Post-Training Data Details}
\label{app:posttraining-data}

\subsection{Post-Training Quality Assessment}
\label{app:posttraining-quality}

Table~\ref{tab:quality_criteria} presents the complete annotation protocol used to train the sample-quality evaluator. Score~0 marks critical defects that invalidate a sample, Score~1 denotes limited but non-fatal issues, and Score~2 denotes a valid sample without the listed defects.

\begin{table}[H]
    \centering
    \caption{Fine-grained quality assessment criteria used for multimodal data filtering.}
    \label{tab:quality_criteria}
    \scriptsize
    \begin{tabularx}{\textwidth}{p{0.18\textwidth} c p{0.20\textwidth} X}
        \toprule
        \textbf{Dimension}        & \textbf{Score} & \textbf{Criterion}                  & \textbf{Description}                                                                                                                                \\
        \midrule
        Image--question alignment & 0              & Non-medical image                   & The image is unrelated to the medical context required by the question.                                                                             \\
                                  & 0              & Modality mismatch                   & The visual modality required by the question is inconsistent with the provided image, e.g., a question referring to CT while the input is an X-ray. \\
                                  & 0              & Anatomical mismatch                 & The organ, anatomical region, or laterality specified by the question is absent from the provided image(s).                                         \\
                                  & 0              & Missing image                       & The question refers to an image or panel that is not provided, or the available images are insufficient for answering the question.                 \\
                                  & 1              & Image-independent question          & The question can be answered without using the visual input.                                                                                        \\
        \midrule
        Information completeness  & 0              & Invalid answer space                & The candidate answers do not contain a valid option or are not mutually compatible with the intended question.                                      \\
                                  & 0              & Insufficient conditions             & Necessary clinical context, parameters, or problem conditions are missing, making the question under-specified or unsolvable.                       \\
                                  & 0              & Incorrect image description         & The textual description or premise is inconsistent with the visual content.                                                                         \\
                                  & 0              & Missed or incorrect visual evidence & Key visual evidence, such as labels or lesions, is incorrectly captured or omitted.                                                                 \\
                                  & 0              & Incorrect answer                    & The answer is incorrect despite sufficient and valid problem conditions.                                                                            \\
                                  & 1              & Missing measurement scale           & The task requires quantitative measurement, but the image does not provide a reliable scale or reference.                                           \\
        \midrule
        Logical validity          & 0              & Ambiguous question                  & The question is ill-posed, internally inconsistent, or insufficiently precise.                                                                      \\
                                  & 0              & Inconsistent response               & The response does not address the question or is logically incompatible with it.                                                                    \\
                                  & 0              & Low-value question                  & The question does not provide meaningful medical or multimodal supervision.                                                                         \\
        \midrule
        Overall validity          & 2              & Valid sample                        & The sample contains sufficient visual and textual evidence, is logically well-formed, and provides a valid target response.                         \\
        \bottomrule
    \end{tabularx}
\end{table}

\subsection{Supervised Fine-Tuning Mixture Composition}
\label{app:sft-mixture}

The final mixture contains 296,725 effective training examples and 241.848M tokens. Vision and text contribute 120.459M and 121.389M tokens, respectively.

\begin{table}[H]
    \centering
    \caption{Composition of the supervised fine-tuning mixture.}
    \label{tab:sft-mixture-composition}
    \small
    \begin{tabular}{l r r r r r}
        \toprule
        \textbf{Category}                 & \textbf{Examples} & \textbf{Tokens} & \textbf{Share} & \textbf{Vision} & \textbf{Text} \\
        \midrule
        General multimodal instruction    & 50,000            & 26.021M         & 10.76\%        & 16.909M         & 9.112M        \\
        Medical visual question answering & 90,734            & 52.644M         & 21.77\%        & 41.699M         & 10.945M       \\
        Medical reasoning data            & 49,808            & 54.868M         & 22.69\%        & 16.469M         & 38.399M       \\
        Stability-aware supervision       & 106,183           & 108.316M        & 44.79\%        & 45.382M         & 62.933M       \\
        \midrule
        Total                             & 296,725           & 241.848M        & 100.00\%       & 120.459M        & 121.389M      \\
        \bottomrule
    \end{tabular}
\end{table}

\subsection{Stability-Aware Routing and Retention}
\label{app:stability-statistics}

Table~\ref{tab:stability-routing} reports the data flow through stability-aware response regeneration. We draw eight candidate responses for each of 176,948 questions. Rule-based answer checking identifies 704,046 correct candidates, and 121,544 questions have at least one correct teacher candidate. This is an intermediate count, not a routing threshold. After stability filtering and final processing, 106,183 examples enter SFT. The selected supervision contributes 108.316M training tokens, including 45.382M visual tokens and 62.933M text tokens.

\begin{table}[H]
    \centering
    \caption{Data flow through stability-aware response regeneration. Percentages for candidates use all generated candidates as the denominator; percentages for questions use the original question set.}
    \label{tab:stability-routing}
    \small
    \renewcommand{\arraystretch}{1.08}
    \begin{tabular*}{\textwidth}{@{\extracolsep{\fill}}lrr@{}}
        \toprule
        \textbf{Stage} & \textbf{Count} & \textbf{Retention} \\
        \midrule
        Input questions & 176,948 & 100.00\% \\
        Generated candidates ($8$ per question) & 1,415,584 & 100.00\% \\
        Candidates passing answer checking & 704,046 & 49.74\% \\
        Questions with at least one valid candidate & 121,544 & 68.69\% \\
        Long-form examples exported after stability filtering & 108,293 & 61.20\% \\
        Final stability-aware SFT examples & 106,183 & 60.01\% \\
        \bottomrule
    \end{tabular*}
\end{table}

The teacher candidates that pass the initial answer check average 418.31 assistant tokens. Of the 704,046 reconstructed answers, 662,503 recover the reference answer (94.10\%). The main paper compares the lengths of the reconstructions and the exported long-form SFT targets.

\section{Training and Implementation Details}
\label{app:training-details}

Table~\ref{tab:training-hyperparameters} consolidates the optimization settings used across continued pretraining and supervised fine-tuning. All CPT and SFT stages use AdamW with $\beta_1=0.9$, $\beta_2=0.95$, $\epsilon=10^{-8}$, and gradient clipping at a maximum norm of 1.0. During warmup--stable CPT and learning-rate decay, we use a weight decay of 0.1 while excluding bias and RMSNorm parameters.

\begin{table}[H]
    \centering
    \caption{Optimization settings for the main training stages. Batch sizes are reported as global and per-device micro-batch sizes.}
    \label{tab:training-hyperparameters}
    \scriptsize
    \setlength{\tabcolsep}{2.5pt}
    \renewcommand{\arraystretch}{1.35}
    \begin{tabular}{@{}>{\centering\arraybackslash}m{0.19\textwidth}>{\centering\arraybackslash}m{0.18\textwidth}>{\centering\arraybackslash}m{0.06\textwidth}>{\centering\arraybackslash}m{0.09\textwidth}>{\centering\arraybackslash}m{0.07\textwidth}>{\centering\arraybackslash}m{0.30\textwidth}@{}}
        \toprule
        \textbf{Stage}            & \textbf{Trainable components} & \textbf{Epochs} & \textbf{Max. length} & \textbf{Batch} & \textbf{Learning-rate schedule}                                                                                                        \\
        \midrule
        Vision-encoder adaptation & Vision encoder, projector     & 1               & 8,192                & 256 / 2        & 3\% warmup to $2\times10^{-6}$, cosine decay to $2\times10^{-7}$                                                                       \\
        Warmup--stable CPT        & All components                & 1               & 12,288               & 256 / 1        & 3\% warmup to $1\times10^{-5}$, then constant                                                                                          \\
        Learning-rate decay       & All components                & 1               & 12,288               & 256 / 1        & Cosine decay from $1\times10^{-5}$ to $2\times10^{-6}$                                                                                 \\
        Supervised fine-tuning    & All components                & 5               & 9,000                & 64 / 2         & 10\% warmup, then cosine decay to zero. Peak $2\times10^{-6}$ for language model and projector and $1\times10^{-6}$ for vision encoder \\
        \bottomrule
\end{tabular}
\end{table}

\subsection{Reinforcement Learning with Verifiable Rewards}
\label{app:rlvr-data}

\subsubsection{Optimization Settings}

RLVR is run for three epochs with a learning rate of $1\times10^{-6}$ and a global batch size of 128. For each prompt, the policy samples eight rollouts at temperature 1. We retain the default KL regularization setting of the training framework. The reward combines format compliance and answer accuracy with a weight ratio of $1{:}9$, placing most of the optimization signal on verifiable correctness while preserving the required response format. Table~\ref{tab:rlvr-hyperparameters} summarizes these settings.

\begin{table}[H]
    \centering
    \caption{Optimization settings for reinforcement learning with verifiable rewards.}
    \label{tab:rlvr-hyperparameters}
    \small
    \renewcommand{\arraystretch}{1.08}
    \begin{tabular*}{\textwidth}{@{\extracolsep{\fill}}lr@{}}
        \toprule
        \textbf{Hyperparameter} & \textbf{Value} \\
        \midrule
        Learning rate & $1\times10^{-6}$ \\
        Global batch size & 128 \\
        Training epochs & 3 \\
        Rollouts per prompt & 8 \\
        Sampling temperature & 1 \\
        Format-to-accuracy reward ratio & $1{:}9$ \\
        \bottomrule
    \end{tabular*}
\end{table}

\subsubsection{Data Composition}

The RLVR resources combine image-grounded clinical questions with text-only medical reasoning. Table~\ref{tab:rlvr-data-composition} reports the provided train and test partitions for each source.

\begin{table}[H]
    \centering
    \caption{Composition of the medical reasoning resources prepared for RLVR.}
    \label{tab:rlvr-data-composition}
    \small
    \renewcommand{\arraystretch}{1.10}
    \begin{tabular*}{\textwidth}{@{\extracolsep{\fill}}c c c c c@{}}
        \toprule
        \textbf{Category} & \textbf{Source} & \textbf{Train} & \textbf{Test} & \textbf{Total} \\
        \midrule
        \multirow{5}{*}{Medical VQA} & PMC-VQA & 7,210 & 790 & 8,000 \\
        & MedSynVQA & 5,371 & 629 & 6,000 \\
        & GMAI-VL-5.5M & 2,721 & 279 & 3,000 \\
        & Internally curated medical data & 1,848 & 206 & 2,054 \\
        & OmniMedVQA + RadImageNet & 426 & 49 & 475 \\
        \midrule
        Text QA & MedQA (English USMLE) & 1,800 & 200 & 2,000 \\
        \bottomrule
\end{tabular*}
\end{table}

\section{Evaluation Details}
\label{app:evaluation-details}

\subsection{Benchmark Overview}

\paragraph{MMMU-Medical-test.} MMMU evaluates expert-level multimodal perception and reasoning with questions collected from examinations, quizzes, and textbooks across six broad disciplines, 30 subjects, and 183 subfields~\citep{yue2024mmmu}. We report the test questions belonging to its Health and Medicine discipline.

\paragraph{MMMU-Pro-Medical-10.} MMMU-Pro strengthens MMMU by removing questions that can be solved reliably without images, expanding the answer space, and introducing a vision-only setting~\citep{yue2025mmmupro}. We use its 10-option Health and Medicine subset, which reduces the influence of option-level shortcuts and random guessing.

\paragraph{MedXpertQA-MM.} MedXpertQA contains 4,460 expert-level medical questions spanning 17 specialties and 11 body systems~\citep{zuo2025medxpertqa}. Its multimodal subset combines diverse medical images with clinical information such as patient records and examination findings, emphasizing medical knowledge and multi-step reasoning rather than caption-level recognition.

\paragraph{PMC-VQA-clean.} PMC-VQA is constructed from biomedical articles and contains approximately 227k visual question-answer pairs associated with 149k images across diverse modalities and diseases~\citep{zhang2023pmcvqa}. We evaluate on the cleaned PMC-VQA split used throughout our experiments.

\paragraph{OmniMedVQA.} OmniMedVQA aggregates authentic medical images from 73 datasets, covering 12 imaging modalities and more than 20 anatomical regions~\citep{hu2024omnimedvqa}. Its broad visual coverage complements the knowledge- and reasoning-oriented benchmarks above.

\subsection{Inference and Prompting Protocol}

All models are evaluated with native thinking disabled (\texttt{enable\_thinking=False}) and a maximum generation length of 4,096 tokens (\texttt{max\_new\_tokens=4096}). MMMU-Medical-test, MMMU-Pro-Medical-10, MedXpertQA-MM, and PMC-VQA-clean use the following reasoning prompt:

\begin{promptbox}
\raggedright\small\ttfamily
You should think step by step and answer with the option's letter from the given choices and put the letter within \string<answer\string> and \string</answer\string>.
\end{promptbox}

OmniMedVQA uses a direct-answer prompt:

\begin{promptbox}
\raggedright\small\ttfamily
Answer the question directly and put the option's letter from the given choices within \string<answer\string> and \string</answer\string>.
\end{promptbox}

The evaluator extracts the option letter enclosed by the answer tags and computes exact-match accuracy. The same benchmark examples, prompts, generation limits, and answer-extraction procedure are used for every evaluated model.

\begin{caseblock}
\subsection{Qualitative Case Studies}
\label{app:qualitative-cases}

Figures~\ref{fig:case-myeloma}--\ref{fig:case-av-node} compare representative responses from the backbone models and InfiMed2 variants. Red text marks an incorrect claim or answer, whereas green text marks correctly identified evidence, reasoning, or conclusions.

\begin{figure}[H]
    \centering
    \casegraphic{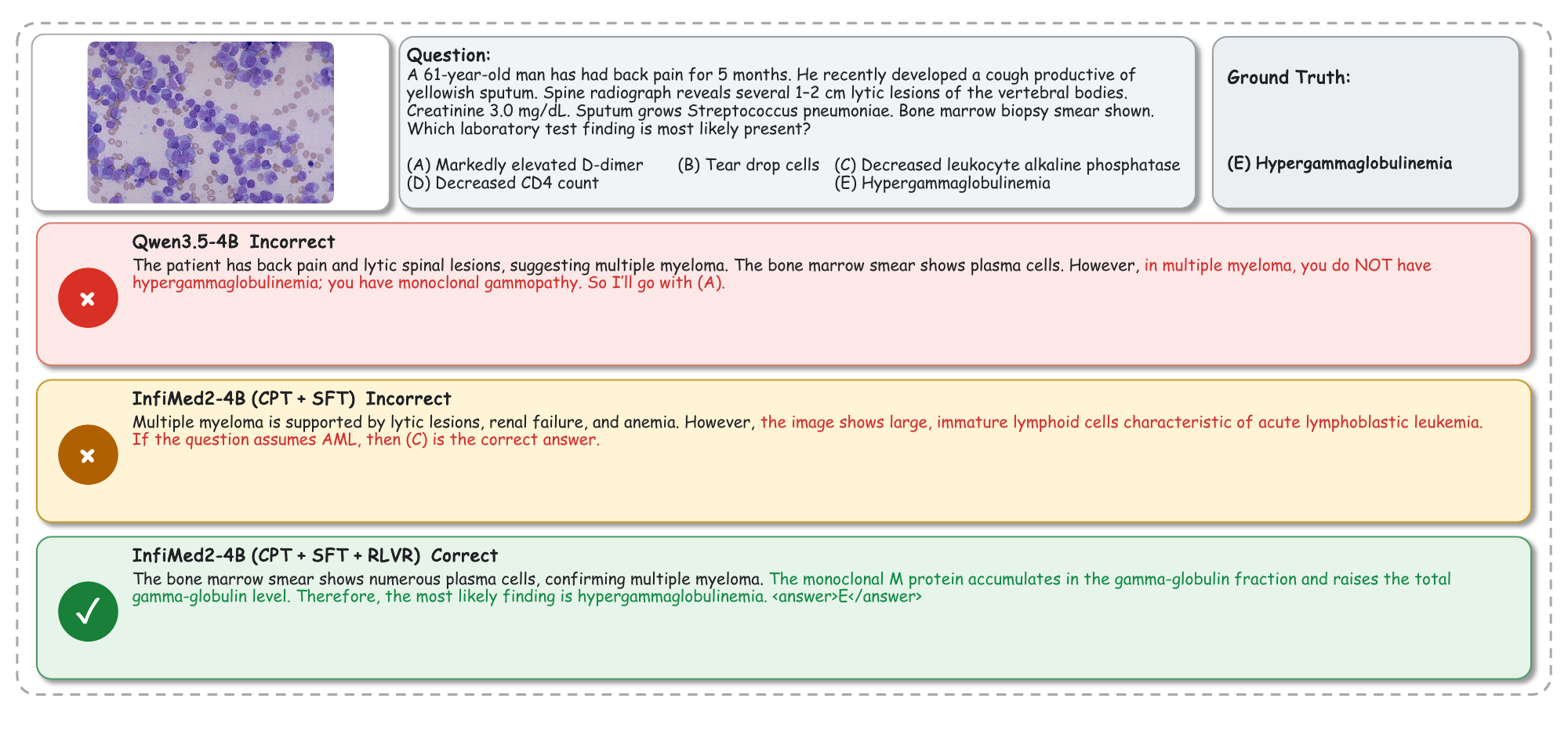}
    \caption{Qualitative comparison on a multiple-myeloma question from MMMU-Medical-test.}
    \label{fig:case-myeloma}
\end{figure}

\noindent\textbf{Case 1: integrating visual and clinical evidence.}
All three responses recover at least part of the clinical pattern, but they differ in how the marrow image is integrated with the accompanying findings. Qwen3.5-4B recognizes multiple myeloma yet incorrectly treats monoclonal gammopathy as incompatible with hypergammaglobulinemia. The CPT+SFT model instead misidentifies plasma cells as lymphoblasts and changes the diagnosis. After RLVR, InfiMed2-4B jointly uses the plasma-cell morphology, osteolytic lesions, renal impairment, and monoclonal protein to select the correct answer. The example illustrates that a correct disease hypothesis is insufficient when the final laboratory implication remains inconsistent with it.
\end{caseblock}

\begin{caseblock}
\begin{figure}[H]
    \centering
    \casegraphic{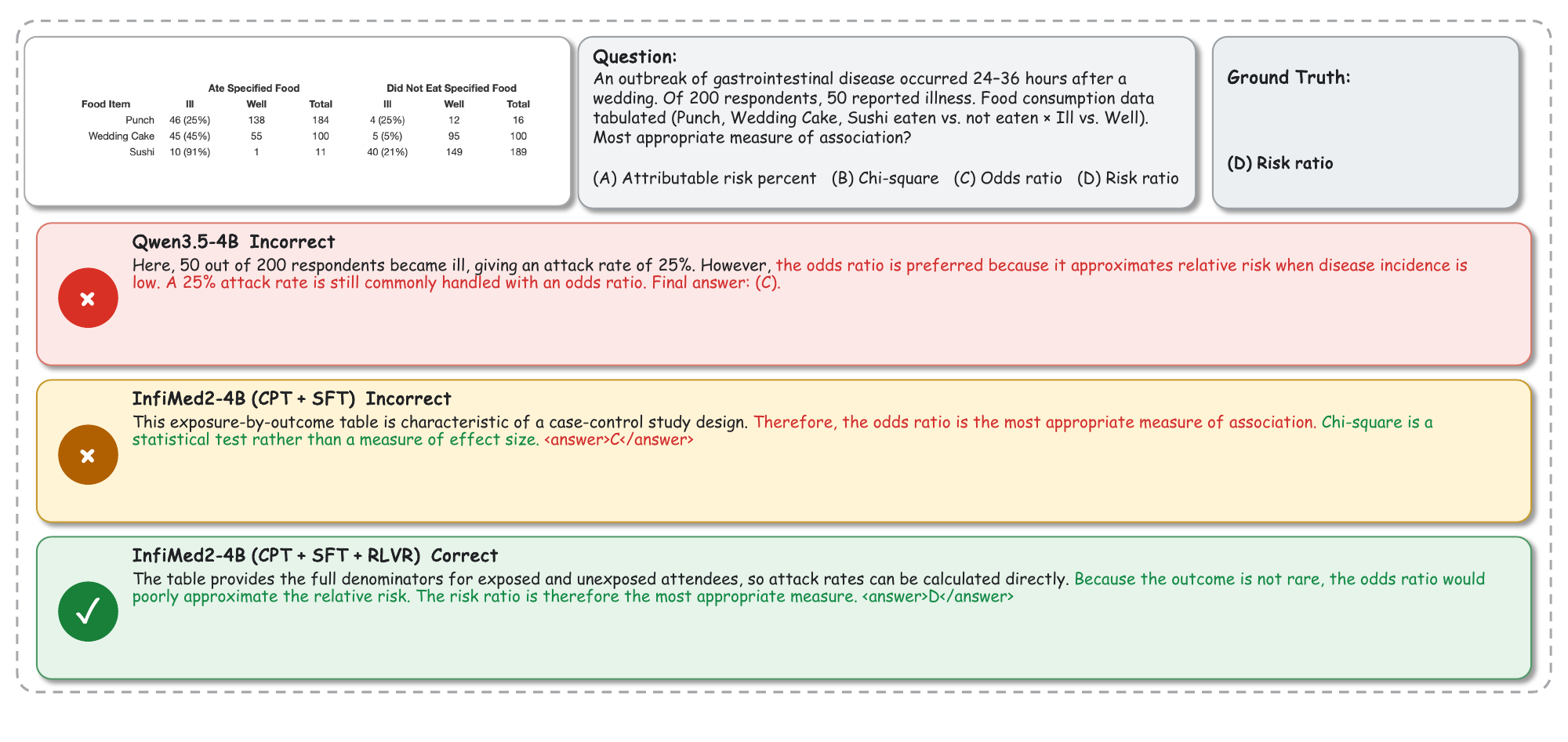}
    \caption{Qualitative comparison on an outbreak-analysis question from MMMU-Medical-test.}
    \label{fig:case-outbreak}
\end{figure}

\noindent\textbf{Case 2: identifying the study design.}
The two incorrect responses focus on the odds ratio, despite the table providing denominators for both exposed and unexposed wedding attendees. One response notices the 25\% attack rate but still applies the rare-disease approximation, while the other misclassifies the retrospective cohort as a case-control study. The RLVR model uses the observable attack rates to identify the risk ratio as the direct measure of association. This case highlights the importance of selecting a statistical quantity from the data-generating design rather than from the superficial appearance of a contingency table.
\end{caseblock}

\begin{caseblock}
\begin{figure}[H]
    \centering
    \casegraphic{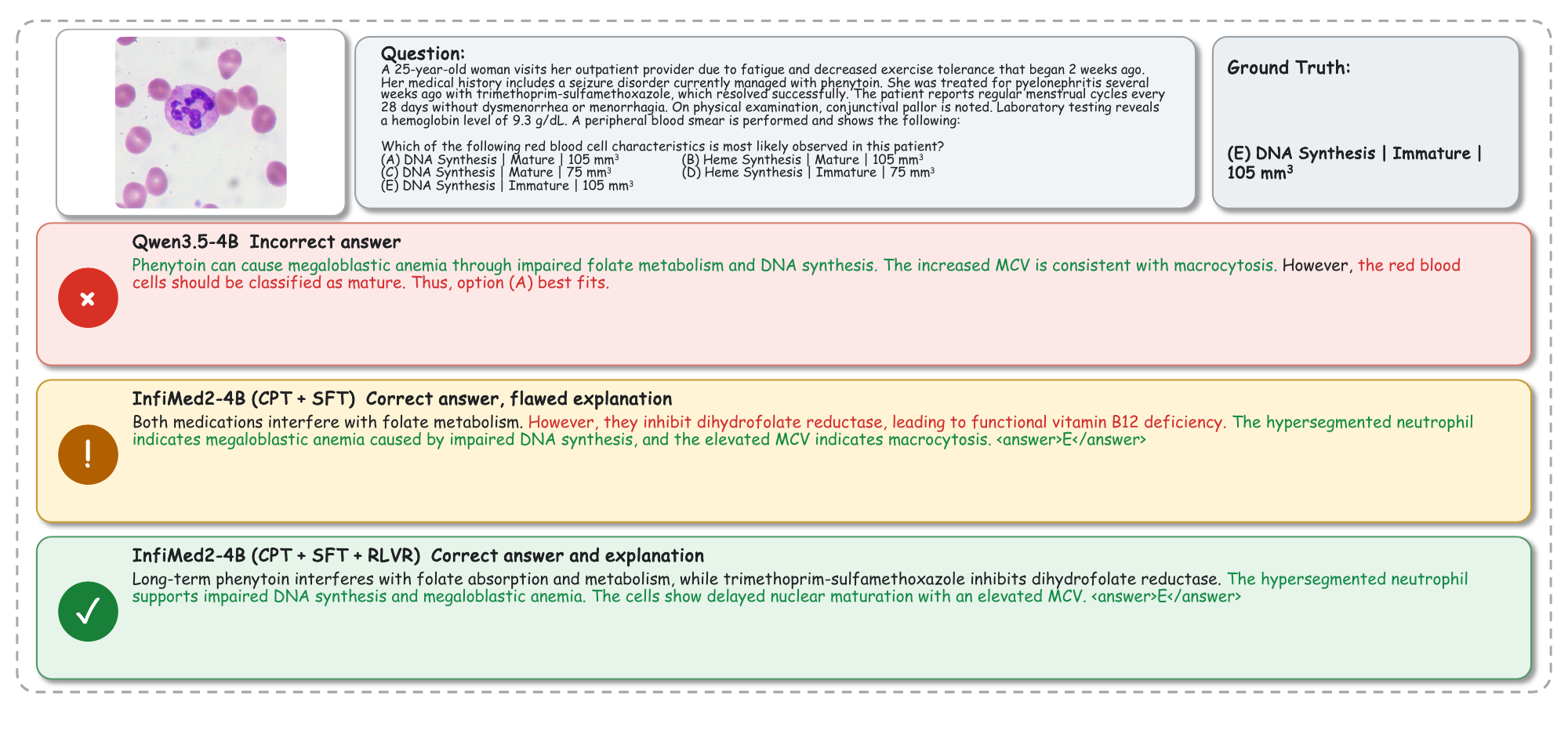}
    \caption{Qualitative comparison on a drug-induced megaloblastic-anemia question from MedXpertQA-MM.}
    \label{fig:case-megaloblastic}
\end{figure}

\noindent\textbf{Case 3: separating answer correctness from explanation quality.}
Qwen3.5-4B correctly associates phenytoin with impaired folate metabolism and macrocytosis but selects the wrong maturation state. The CPT+SFT model reaches the correct option while incorrectly attributing the mechanism to functional vitamin B12 deficiency and treating both drugs as dihydrofolate-reductase inhibitors. In contrast, the RLVR response distinguishes phenytoin-related folate disruption from the action of trimethoprim-sulfamethoxazole and aligns the mechanism, smear finding, and selected option. This example motivates evaluating the factual consistency of an explanation even when its final answer is correct.
\end{caseblock}

\begin{caseblock}
\begin{figure}[H]
    \centering
    \casegraphic{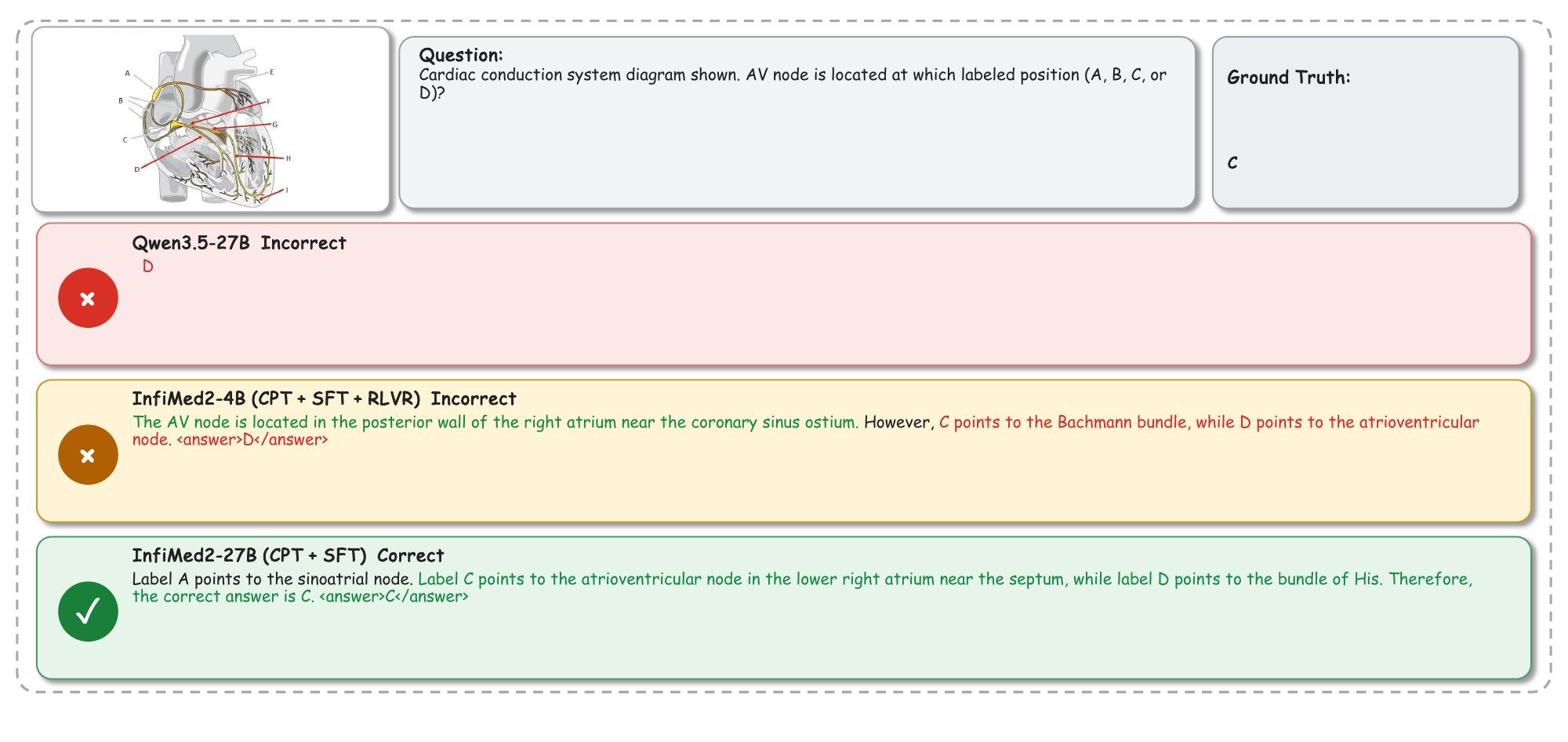}
    \caption{Qualitative comparison on cardiac-conduction anatomy from MMMU-Medical-test.}
    \label{fig:case-av-node}
\end{figure}

\noindent\textbf{Case 4: grounding anatomical knowledge in diagram labels.}
Qwen3.5-27B provides only the incorrect label. InfiMed2-4B with RLVR accurately describes the anatomical location of the atrioventricular node but maps that description to the wrong marker in the diagram. InfiMed2-27B correctly distinguishes the sinoatrial node, atrioventricular node, and bundle of His and maps the atrioventricular node to label~C. The comparison exposes a residual failure mode in which verbal anatomical knowledge is correct but its spatial grounding is not, while also showing that the larger InfiMed2 variant resolves the visual-label correspondence in this example.
\end{caseblock}

% Additional appendix material can be enabled when the required records become
% available: curation prompts and label taxonomies, RLVR hyperparameters,
% benchmark prompt templates and answer extraction, qualitative examples, and
% data provenance and licensing notes.

\end{document}